\documentclass[review]{elsarticle}

\usepackage{hyperref}
\usepackage{subfig}
\usepackage{lineno}
\usepackage{amsmath}
\DeclareMathOperator*{\argminA}{arg\,min}

\usepackage[linesnumbered,ruled]{algorithm2e}
\SetKwRepeat{Do}{do}{while}
\begin{document}

\begin{frontmatter}

\title{A Novel Disparity Transformation\\ Algorithm for Road Segmentation}

\author{Rui Fan$^{a*}$, Mohammud Junaid Bocus$^a$, Naim Dahnoun$^b$}
\address[mymainaddress]{Visual Information Laboratory, the University of Bristol, BS8 1UB, UK.}
\address[mysecondaryaddress]{Department of Electrical and Electronic Engineering, the University of Bristol, BS8 1UB, UK.
}

\begin{abstract}
The disparity information provided by stereo cameras has enabled advanced driver assistance systems to estimate road area more accurately and effectively. In this paper, a novel disparity transformation algorithm is proposed to extract road areas from dense disparity maps by making the disparity value of the road pixels become similar. The transformation is achieved using two parameters: roll angle $\gamma$ and  fitted disparity value $d$ with respect to each row. To achieve a better processing efficiency, golden section search and dynamic programming are utilised to estimate $\gamma$ and $d$, respectively. By performing a rotation around $\gamma$, the disparity distribution of each row becomes very compact. This further improves the accuracy of the road model estimation, as demonstrated by the various experimental results in this paper. Finally, the Otsu's thresholding method is applied to the transformed disparity map and the roads can be accurately segmented at pixel level. 
\end{abstract}


\end{frontmatter}


\section{Introduction}
\label{sec.pt_introduction}
\begin{figure}[!b]
	\centering
	\includegraphics[width=0.60\textwidth]{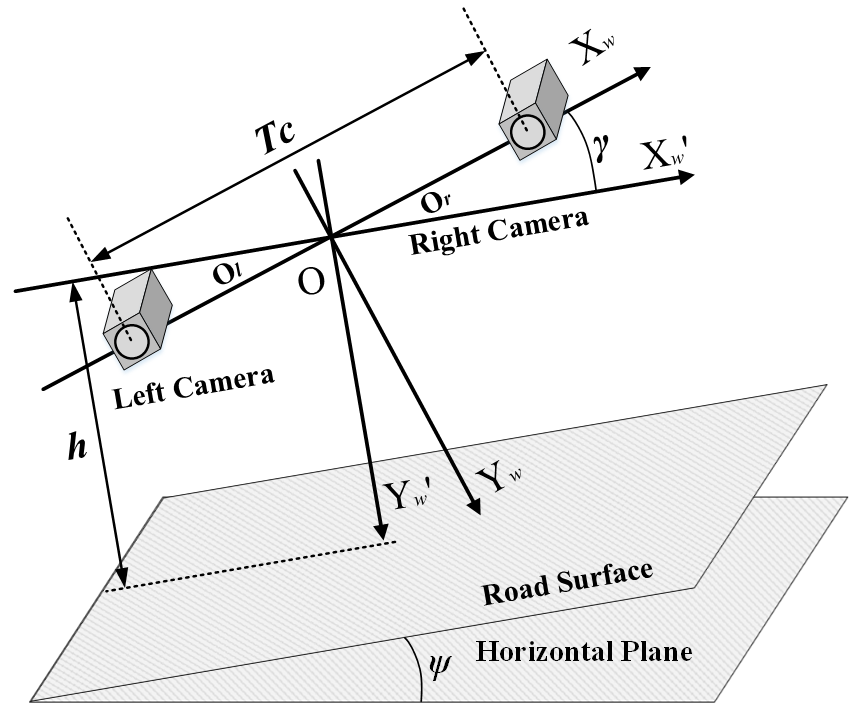}
	\caption{An example of roll angle $\gamma$ and road bank angle $\psi$. $T_c$ and $h$ represent the baseline and the height of the stereo rig, respectively. }
	\label{fig.roll_bank_angle}
\end{figure}

\begin{figure}[!b]
	\centering
	\subfloat[dense disparity map $\ell^{ori}$ ($\gamma\neq0$).]{
		\includegraphics[width=0.488\textwidth]{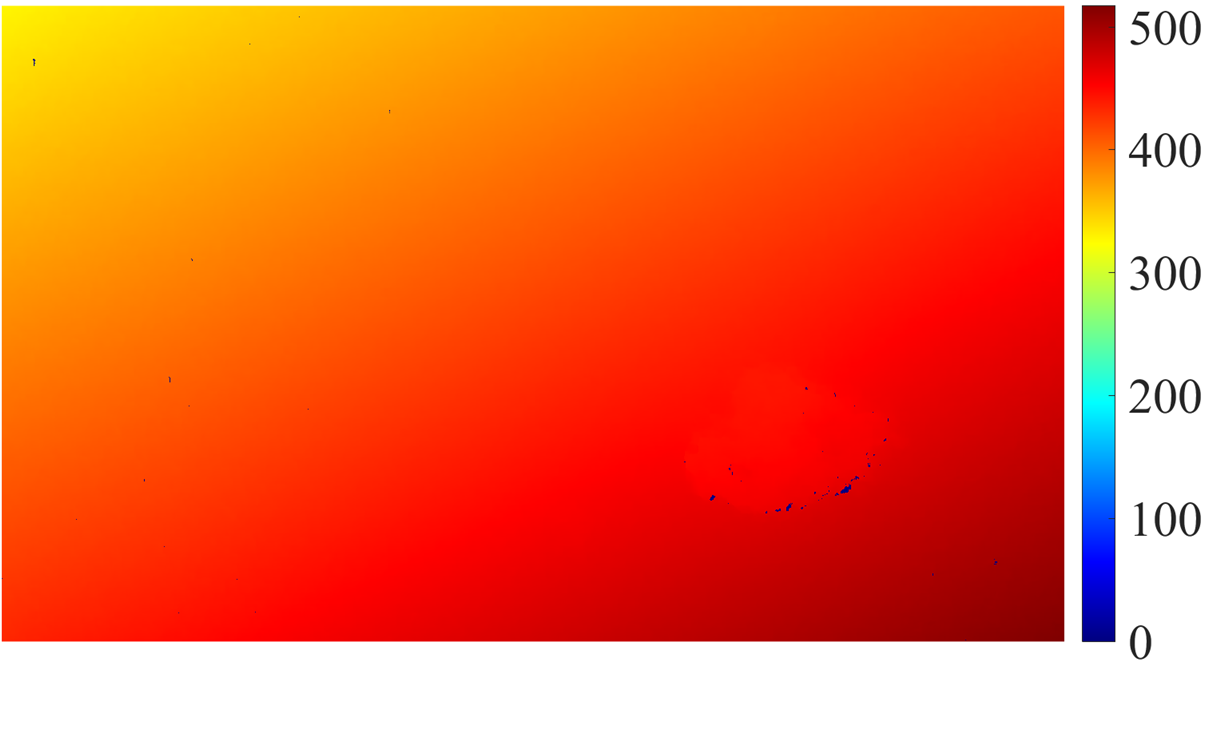}
		\label{fig.disp_map1}
	}
	\quad
	\subfloat[ v-disparity map.]{
		\includegraphics[width=0.230\textwidth]{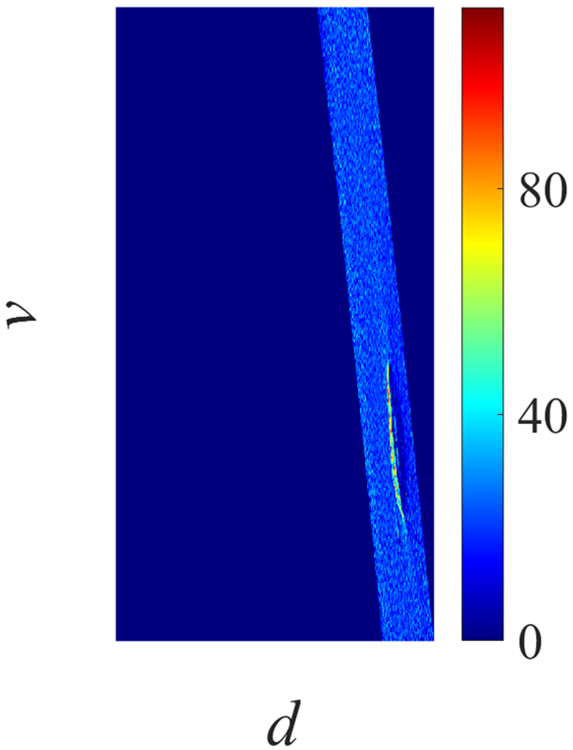}
		\label{fig.v_disp_map1}
	}
	\caption{The input dense disparity map ($\gamma\neq0$) and its corresponding v-disparity map.}
	\label{fig.unwell_roll_angle}
\end{figure}

3-D information is very important for various intelligent transportation systems (ITSs), such as lane detection \cite{Fan2016}, road condition assessment \cite{Fan2018c}, free space estimation \cite{Badino2007} and visual simultaneous localisation and mapping (SLAM) \cite{Newman2006}. Computer stereo vision is one of the most commonly used techniques in ITSs for dense 3-D data acquisition. The latter is achieved by estimating the relative positional difference between each pair of corresponding points in the left and right images.  This difference is generally known as disparity \cite{Fan2018, Fan2018a}. Since Labayrade et al. proposed the concept of \textquoteleft v-disparity map' in 2002, a lot of research has been carried out to extract road areas from dense disparity maps \cite{Labayrade2002}. These algorithms generally fit a linear or quadratic road model to the best path in the v-disparity map and then analyse the difference between the actual and fitted disparity values to determine the road pixels \cite{Fan2018d}. However, the non-zero roll angle $\gamma$ (see Figure \ref{fig.roll_bank_angle}) usually makes the best path in the v-disparity map ambiguous (see Figure \ref{fig.v_disp_map1}), which further severely affects the accuracy of the road model estimation \cite{Evans2018}. Therefore, $\gamma$ has to be estimated beforehand to address the above issue. 

Over the past decade, various technologies, e.g., inertial measurement units (IMU) \cite{Ryu2002, Ryu2004, EricTseng2007, Oh2013} and passive sensing \cite{Schlipsing2011, Evans2018}, have been utilised to estimate $\gamma$. The methods based on IMU usually combine the data obtained from different sensors, e.g., GPS, accelerometers and gyroscopes, to provide an accurate estimation of the vehicle state \cite{EricTseng2007, Oh2013}. This not only makes the set-up cost unreasonably high but also introduces additional complexity in the estimation procedure  \cite{Schlipsing2011}. Furthermore,  the bank angle $\psi$ (see Figure \ref{fig.roll_bank_angle}) is always assumed to be zero in these approaches and only $\gamma$ is considered in the estimation process. However, in real-world applications, $\psi$ is not always zero and hence both angles have to be estimated independently although this is not usually a straightforward process \cite{Ryu2004}. In recent years, some passive sensor-based methods \cite{Ozgunalp2017, Fan2018, Schlipsing2011, Schlipsing2012,Labayrade2003, Skulimowski} have been proposed to estimate $\gamma$. Most of them used a single camera to acquire road data and some ideal assumptions, such as constant lane width, were made to ensure these algorithms work well \cite{Labayrade2003}. 
However, the roll angle usually changes over time in actual cases, and therefore such assumptions are not always valid. In this regard, some authors resorted to 3-D information in order to estimate roll angle more accurately \cite{Ozgunalp2017, Evans2018, Labayrade2003, Skulimowski}. In \cite{Labayrade2003} and \cite{Skulimowski}, the authors assumed that the road surface is a ground plane and estimated the roll and pitch angles from a v-disparity map. In \cite{Ozgunalp2017, Fan2018, Evans2018}, a ground plane $d(u,v)=\gamma_0+\gamma_1u +  \gamma_2v $ (where $(u,v)$ represents the coordinate of a pixel in the disparity map) is fitted to a small patch which is selected from the near field in the disparity map and $\gamma=\arctan(-\gamma_1/\gamma_2)$. However, the above stereo vision-based algorithms are only suitable for a flat road surface which can be assumed to be a ground plane \cite{Fan2018}.
Furthermore, selecting a proper patch for plane fitting is  always a challenging task because it may contain an obstacle or a pothole which can severely affect the fitting accuracy \cite{Evans2018}. 
Hence in this paper, we first propose a highly robust roll angle estimation algorithm which is suitable for both flat and non-flat roads. By performing a rotation around $\gamma$, the disparity distribution of each row become very compact and this further improves the accuracy of the road model estimation \cite{Evans2018}. Then, the dense disparity map is transformed using two parameters $\gamma$ and $d$, where $d$ denotes the fitted disparity value with respect to each row $v$ and it can be computed using our previously published algorithm in \cite{Ozgunalp2017}. In the transformed disparity map, the disparity value of the road pixels become very similar, which makes the extraction of road area much easier.

The proposed algorithm is composed of three main steps: $\gamma$ estimation, $d$ estimation and disparity transformation. In this paper, the vertical road profile is formulated as a parabola $d(v)=\alpha_0 +\alpha_1 v+ \alpha_2 v^2$. For the stereo rig which is perfectly horizontal to the road surface, the actual disparity values $\tilde{d}$ on each row $v$ are very similar but they reduce gradually when $v$ decreases. Also, the difference between the actual and fitted disparity values, i.e., $\tilde{d}$ and $d$, is close to zero. However, when $\gamma$ is non-zero, the actual disparity values $\tilde{d}$ on each row $v$ will change gradually (see Figure \ref{fig.disp_map1}) and the disparity distribution of each row spreads out (see Figure \ref{fig.v_disp_map1}). This severely affects the accuracy of the road model estimation and thus should be rectified beforehand. Firstly, the disparity map is rotated around different angles. For each rotation, we fit $d(v)$ using least squares fitting (LSF) and obtain the corresponding minimum energy $E_{min}$ (see Figure \ref{fig.gamma_curve}). $\gamma$ can then be estimated by finding the local minima of the curve in Figure \ref{fig.gamma_curve}. To further improve the processing efficiency, we use golden section search (GSS) to reduce the search range, and thus the roll angle $\gamma$ can be estimated more efficiently. In the second step, we use our previously published algorithm \cite{Ozgunalp2017, Fan2018d} to estimate $d$, where the best path is extracted by optimising the v-disparity map using dynamic programming (DP). Finally, the original disparity map is transformed according to the values of $\gamma$ and $d$, which makes the road area very distinguishable. 

The remainder of this paper is structured as follows: section \ref{sec.pd_ad} describes the proposed algorithm. In section \ref{sec.pd_er}, the experimental results are illustrated and the performance of the proposed algorithm is evaluated. Finally, section \ref{sec.pd_conclusion} concludes the paper. 

\section{Algorithm Description}
\label{sec.pd_ad}

\subsection{$\gamma$ Estimation}
\label{sec.gamma_tran}

Roll angle is an important parameter in vehicle control systems (VCSs) because it is commonly used to estimate vehicle states and maintain the stability of the VCSs \cite{EricTseng2007}. In this paper, the input dense disparity map $\ell^{ori}$ is estimated using our previously published algorithm \cite{Fan2017, Fan2018}. The parameter vector $\boldsymbol{\alpha}=[\alpha_0,\alpha_1,\alpha_2]^\top$ can be estimated by solving a least squares problem as follows:

\begin{equation}
\boldsymbol{\alpha}={\argminA_{\boldsymbol{\alpha}}}\ E
\label{eq.estimate_road_model}
\end{equation}
where
\begin{equation}
E=\sqrt{\frac{1}{n+1}\sum_{i=0}^{n}
	\big(\tilde{d}_i-\big(\alpha_0+\alpha_1v_i+\alpha_2{v_i}^2\big)\big)^2}
\label{eq.road_energy}
\end{equation}

\begin{figure}[!b]
  	\centering
  	\includegraphics[width=0.76\textwidth]{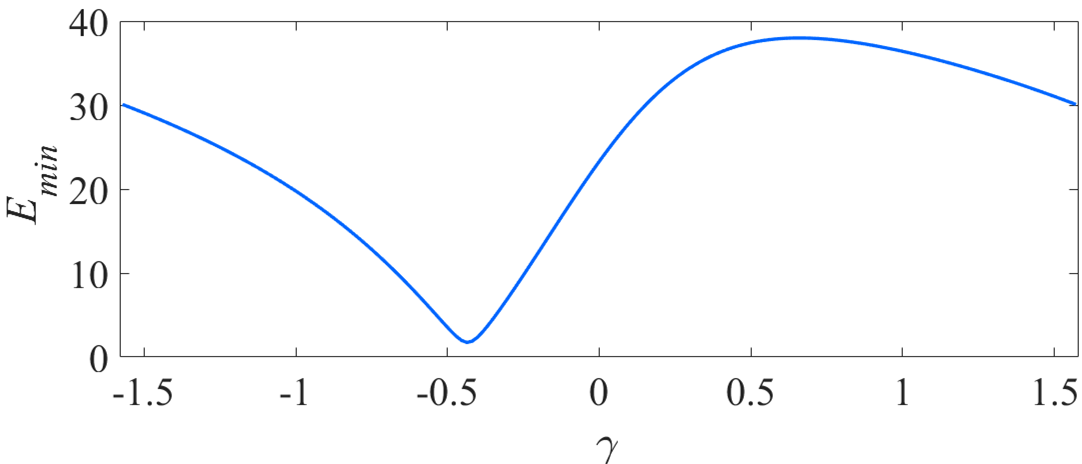}
  	\caption{The relationship between the minimum energy $E_{min}$ and different angles $\gamma$. }
  	\label{fig.gamma_curve}
\end{figure}

\begin{figure}[!t]
	\centering
	\subfloat[rotated dense disparity map $\ell^{rot}$ ($\gamma=0$).]{
		\includegraphics[width=0.56\textwidth]{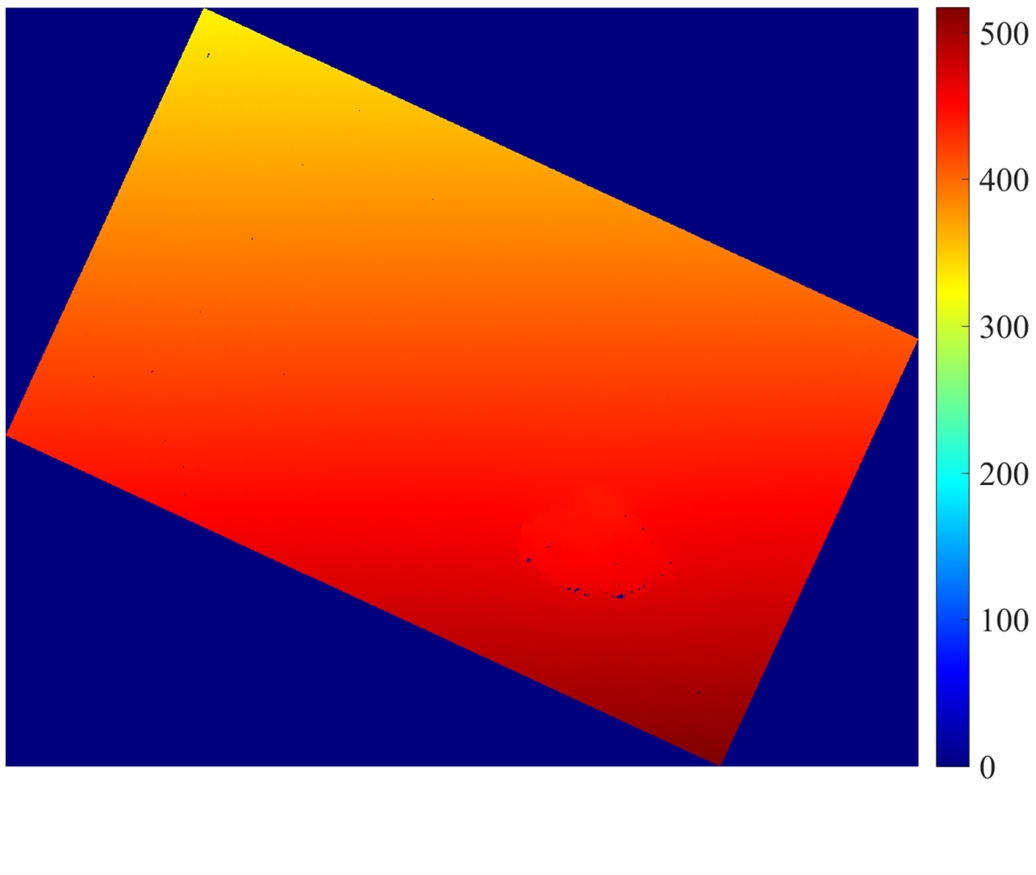}
		\label{fig.disp_map2}
	}
	\subfloat[ v-disparity map.]{
		\includegraphics[width=0.272\textwidth]{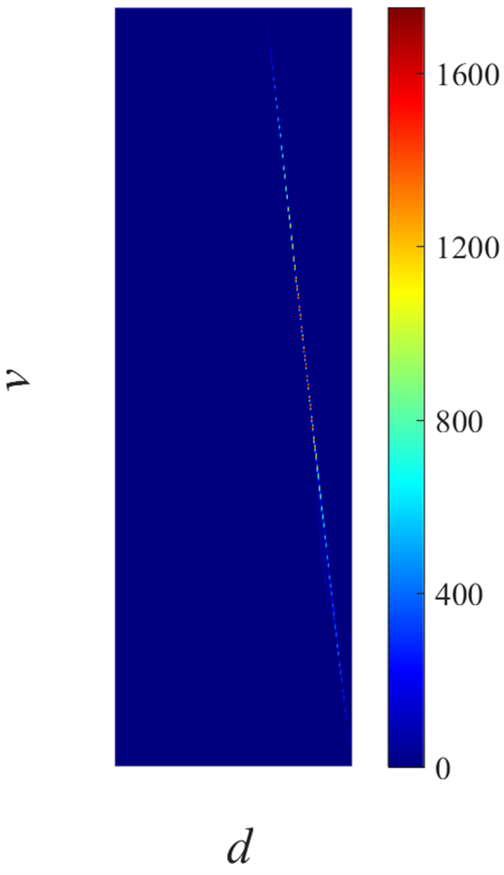}
		\label{fig.v_disp_map2}
	}
	\caption{The rotated dense subpixel disparity map and its corresponding v-disparity map.}
\end{figure}

For a stereo rig which is ideally horizontal to the road surface, the roll angle of the binocular system is zero. The disparity values for each  row are similar  while they change gradually in the vertical direction (see Figure \ref{fig.disp_map2}). However, a non-zero roll angle introduced from the set-up installation makes the disparity values change gradually in each row (see Figure \ref{fig.disp_map1}), and therefore the disparity distribution of each row becomes less compact (see Figure \ref{fig.v_disp_map1}) compared to the case when $\gamma$ is zero (see Figure \ref{fig.v_disp_map2}). This greatly affects the accuracy of the LSF and makes $E_{min}$ higher than the desired value. Therefore, the main idea of the proposed roll angle estimation algorithm is to rotate $\ell^{ori}$ at different angles and then find the angle at which the minimum $E_{min}$ is obtained. An example of the relationship between $E_{min}$ and $\gamma$ is shown in Figure \ref{fig.gamma_curve}.

To rotate $\ell^{ori}$ around a given angle $\gamma$, each coordinate $(u,v)$ in the original disparity map is transformed to a new coordinate $(u',v')$ using Eq. \ref{eq.u_transform} and \ref{eq.v_transform}. where $(u_o,v_o)$ is the coordinate of the centre of $\ell^{ori}$.

\begin{equation}
u{'}=(u-u_o)\cos\gamma+(v-v_o)\sin\gamma
\label{eq.u_transform}
\end{equation}
\begin{equation}
v{'}=(v-v_o)\cos\gamma-(u-u_o)\sin\gamma
\label{eq.v_transform}
\end{equation}

After the coordinate translation and rotation, the position $(u',v')$ in the rotated disparity map $\ell^{rot}$ has the same disparity value as the position $(u,v)$ in $\ell^{ori}$, and the energy function in Eq. \ref{eq.road_energy} can thus be rewritten as Eq. \ref{eq.road_energy_new}.

\begin{equation}
E=\sqrt{\frac{1}{n+1}\sum_{i=0}^{n}
	\big(\tilde{d}_i-\big(\alpha_0+\alpha_1v'_i+\alpha_2{v'_i}^2\big)\big)^2}
\label{eq.road_energy_new}
\end{equation}

\begin{algorithm}[t!]
	\KwData{$\ell^{ori}$}
	\KwResult{$\gamma$}
	set $\gamma_1$ and $\gamma_2$ to $-\pi/2$ and $\pi/2$, respectively\;
	minimise $E$ in Eq. \ref{eq.road_energy_new} with respect to $\gamma_1$ and get $E_{min1}$\;
	minimise $E$ in Eq. \ref{eq.road_energy_new} with respect to $\gamma_2$ and get $E_{min2}$\;
	\While{$\gamma_2-\gamma_1> \varepsilon$}{
		set $\gamma_3$ and $\gamma_4$ to $k\gamma_1+(1-k)\gamma_2$ and $k\gamma_2+(1-k)\gamma_1$, respectively\;
		minimise $E$ in Eq. \ref{eq.road_energy_new} with respect to $\gamma_3$ and get $E_{min3}$\;
		minimise $E$ in Eq. \ref{eq.road_energy_new} with respect to $\gamma_4$ and get $E_{min4}$\;
		\uIf{$E_{min3}>E_{min4}$}{
			$\gamma_1$ is replaced by $\gamma_3$\;}
		\Else{$\gamma_2$ is replaced by $\gamma_4$\;}{
		}
	}
	\caption{$\gamma$ estimation using GSS.}
	\label{al.gamma_estimation}
\end{algorithm}

An arbitrary $v'$ can be computed from $u$ and $v$ using Eq. \ref{eq.v_transform} and the corresponding $E_{min}$ is obtained by solving the energy minimisation problem in Eq. \ref{eq.estimate_road_model} using the LSF. It is to be noted that whether the disparity map is rotated around $\gamma$ or $\gamma+\pi$, the same $E_{min}$ will be obtained because $\cos(\gamma+\pi)=-\cos\gamma$ and $\sin(\gamma+\pi)=-\sin\gamma$. Therefore, we set the interval of $\gamma$  to $(-\pi/2,\pi/2]$ and find the desirable roll angle which corresponds to the local minima as shown in Figure \ref{fig.gamma_curve}.

However, finding the local minima is a computationally intensive task because we have to go through the whole interval. Furthermore, in order to obtain an accurate $\gamma$ which corresponds to the minimum $E_{min}$, the step size $ \varepsilon$ should be set to a very small and practical
value. 
Hence in this paper, GSS is utilised to reduce the search range within the interval $(-\pi/2,\pi/2]$. 
The procedures of the proposed $\gamma$ estimation algorithm are given in algorithm \ref{al.gamma_estimation}, where $k=0.618$ represents the golden section factor. More details on the GSS are available in \cite{Pedregal2006}. 

The dense disparity map is rotated around $\gamma$ which is estimated using algorithm \ref{al.gamma_estimation}. The rotated disparity map is illustrated in Figure \ref{fig.disp_map2}.
Then, we create the corresponding v-disparity map (see Figure \ref{fig.v_disp_map2}), where we can observe that the disparity distribution of each row becomes very compact. The performance evaluation of the proposed $\gamma$ estimation algorithm will be discussed in Section \ref{sec.pd_er}.

\subsection{$d$ Estimation}
\label{sec.d_est}

\begin{algorithm}[b!]
	\SetKwInOut{Input}{Input}
	\SetKwInOut{Output}{Output}
	\Input{optimal solution $\boldsymbol{M}=[\boldsymbol{d}, \boldsymbol{v}]^\top$}
	\Output{$\boldsymbol{\alpha}$}
	\For{$iteration$ $\gets$ $1$ \text{to} $N$}{
		randomly select a set of candidates from $\boldsymbol{M}$\;
		fit a parabola to the selected candidates and get $\boldsymbol{\alpha}$\; 
		determine the inliers $\mathcal{I}$ and outliers $\mathcal{O}$\;
		compute the ratio  $\eta=n_{\mathcal{I}}/(n_{\mathcal{I}}+n_{\mathcal{O}})$\;
		record the values of $\boldsymbol{\alpha}$ and $\eta$\;
	}
	use the parameters of $\boldsymbol{\alpha}$ which correspond to the smallest $\eta$\;
	\Do{$n_{\mathcal{O}}\neq\emptyset$}
	{
		determine the inliers $\mathcal{I}$ and outliers $\mathcal{O}$\;
		remove $\mathcal{O}$ from $\boldsymbol{M}$\;
		fit a parabola to $\boldsymbol{M}$ and update $\boldsymbol{\alpha}$\;
		
	}
	\caption{$\boldsymbol{\alpha}$ estimation optimised with RANSAC.}
	\label{al.dp_lsf}
\end{algorithm}

The disparity projection of the road pixels on the v-disparity map can be extracted using DP by searching for every possible solution and selecting the one $\boldsymbol{M}=[\boldsymbol{d},\boldsymbol{v}]^\top$ with the minimum energy, where $\boldsymbol{d}=[d_0,d_1,\cdots,d_m]^\top$ and $\boldsymbol{v}=[v_0,v_1,\cdots,v_m]^\top$ record the path of the optimal solution. More details on DP are provided in our previous work \cite{Ozgunalp2017}. In this section, we mainly discuss the optimisation of $\boldsymbol{\alpha}$ estimation where Random Sample Consensus (RANSAC) is utilised to reduce the effects caused by outliers. The steps of estimating $\boldsymbol{\alpha}$ are given in algorithm \ref{al.dp_lsf}. The RANSAC iterates $N$ times, where $N$ is empirically set to $20$ in this paper. For each iteration, we select a set of $t$ candidates $[d_j,v_j]^\top$ from the optimal solution $\boldsymbol{M}$ and interpolate them into a parabola. In this way the corresponding parameter vector $\boldsymbol{\alpha}=[\alpha_0, \alpha_1, \alpha_2]^\top$ can be estimated and the number of inliers $\mathcal{I}$ and outliers $\mathcal{O}$ can be computed. We record the parameters of $\boldsymbol{\alpha}$ and the corresponding ratio $\eta=n_{\mathcal{I}}/(n_{\mathcal{I}}+n_{\mathcal{O}})$ for each iteration, where $n_{\mathcal{I}}$ and $n_{\mathcal{O}}$ denote the number of inliers and outliers, respectively. Then, we select the parameters of $\boldsymbol{\alpha}$ which correspond to the smallest ratio $\eta$.
Finally, $\mathcal{O}$ are removed from $\boldsymbol{M}$ and the parameters of $\boldsymbol{\alpha}$ are updated iteratively by fitting a parabola to the updated $\boldsymbol{M}$. The iterations continue until the number of outliers reaches zero. The fitted disparity for each row $v$ can thus be computed as: $d(v)=\alpha_0+\alpha_1 v+\alpha_2 v^2$.

\subsection{Disparity Transformation and Road Segmentation}
\label{sec.abet}

\begin{algorithm}[b!]
	\SetKwInOut{Input}{Input}
	\SetKwInOut{Output}{Output}
	\Input{$\ell^{rot}$, $\gamma$ and $d(v)$}
	\Output{$\ell^{trf}$}
	update each disparity value in $\ell^{rot}$ with $\tilde{d}-d(v)+\delta$\;
	rotate the updated $\ell^{rot}$ around $-\gamma$\;
	\caption{Disparity map transformation. }
	\label{al.qbet}
\end{algorithm}

\begin{figure}[!t]
	\centering
	\subfloat[transformed dense disparity map $\ell^{trf}$.]{
		\includegraphics[width=0.48\textwidth]{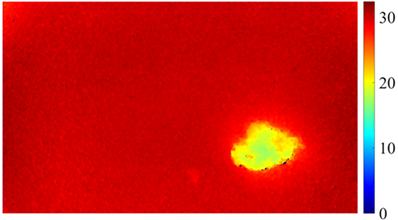}
		\label{fig.disp_map4}
	}
	\subfloat[road segmentation. The area in violet is the extracted road area. ]{
	\includegraphics[width=0.48\textwidth]{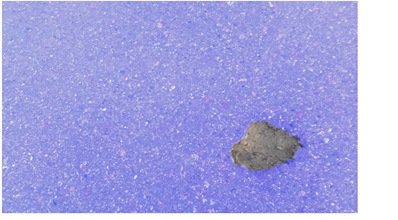}
	\label{fig.left}
}
	\caption{Disparity transformation and road segmentation.}
\end{figure}

Using the values of $\gamma$ and $d$ estimated in subsections \ref{sec.gamma_tran} and \ref{sec.d_est}, the disparity map in Figure \ref{fig.disp_map1} can now be transformed. Firstly, each disparity value $\tilde{d}$ in the rotated disparity map is updated with $\tilde{d}-d(v)+\delta$, where $\delta$ is set to $30$ in this paper to ensure that the disparity values in the updated  $\ell^{rot}$ are non-negative. The updated $\ell^{rot}$ is then rotated around $-\gamma$ to get the transformed disparity map $\ell^{trf}$. More details on the proposed disparity transformation method are given in algorithm \ref{al.qbet}. The corresponding result is shown in Figure \ref{fig.disp_map1}, where we can clearly see that the disparity value of the road pixels becomes very similar but they differs significantly from those of the pothole pixels. Then, we perform Otsu's thresholding method on the transformed disparity map and the road area can be easily extracted. The corresponding  road segmentation result is shown in Figure \ref{fig.left}, where the area in violet is the extracted road surface.

\section{Experimental Results}
\label{sec.pd_er}

We first analyse the computational complexity of the proposed $\gamma$ estimation algorithm. The computational complexity for estimating the roll angle is $O(\frac{\pi}{\varepsilon})$, where $\varepsilon$ is the step size. The GSS reduces the search range exponentially as the interval size becomes only $k^n\pi$ after the $n$th iteration. This reduces the computational complexity to $O(\log_{k}\frac{\varepsilon}{\pi})$. In our experiments, $\varepsilon$ is set to $\pi/1800$ (approximately $0.1^\circ$), and the GSS-based $\gamma$ estimation algorithm only needs to iterate $16$ times to get the desirable roll angle with a precision of $\pm0.1^\circ$. The proposed algorithm is implemented in MATLAB 2018a platform on an Intel Core i7-8700k CPU (3.7GHz) using a single thread. The runtime for a single frame with a resolution of $1249\times610$ is 1.62 seconds. Although the proposed algorithm does not run in real time, we believe that its speed can be increased in the future by exploiting the parallel computing architectures. 

\begin{figure}[!t]
	\centering
	\subfloat[without Gassian noise.]{
		\includegraphics[width=0.450\textwidth]{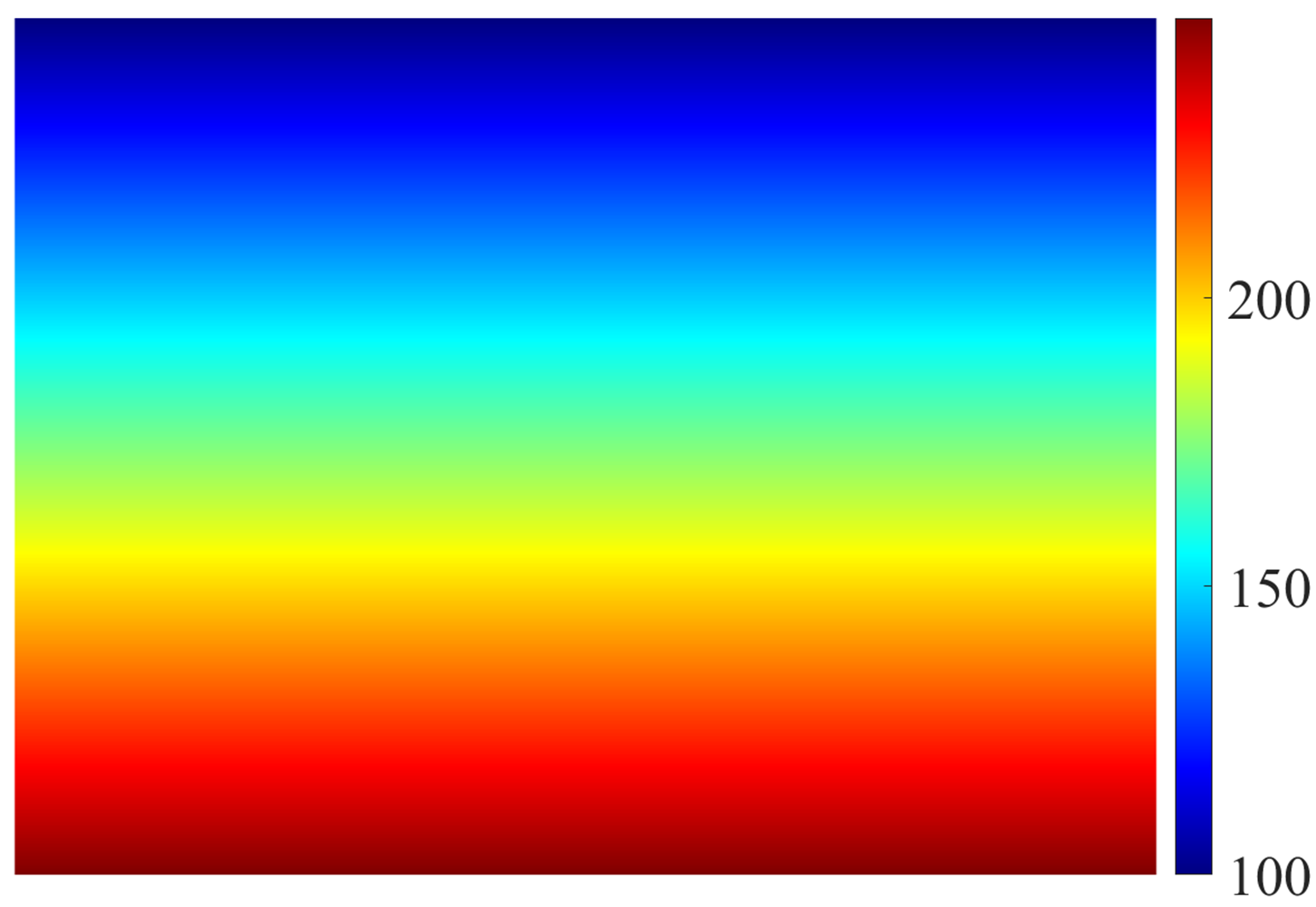}
		\label{fig.disp_gt1}
	}
	\subfloat[with Gaussian noise, $\kappa=50$.]{
		\includegraphics[width=0.450\textwidth]{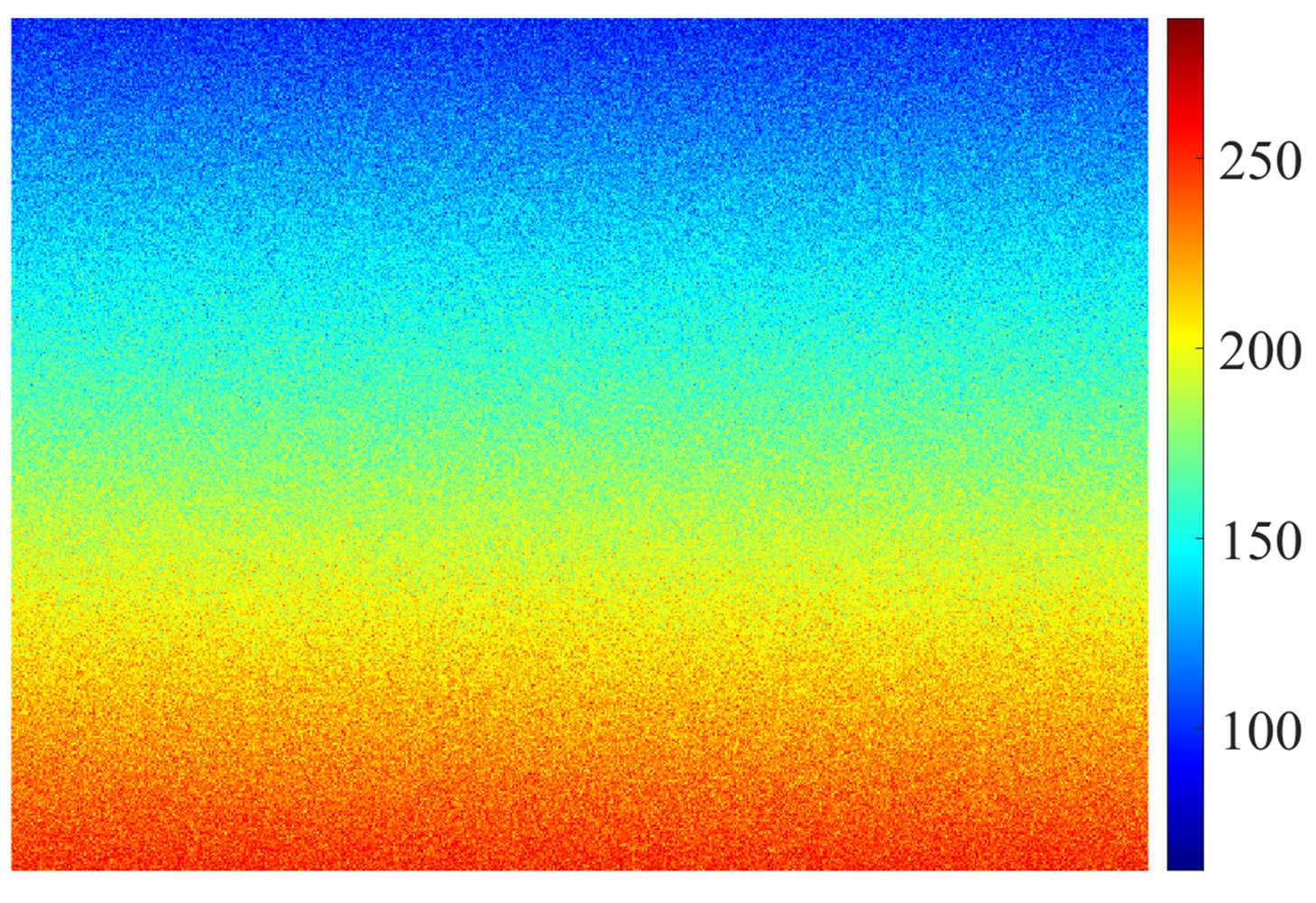}
		\label{fig.disp_gt2}
	}
	\caption{Ground truth disparity maps. }
\end{figure}

\begin{figure}[!b]
	\centering
	\includegraphics[width=1\textwidth]{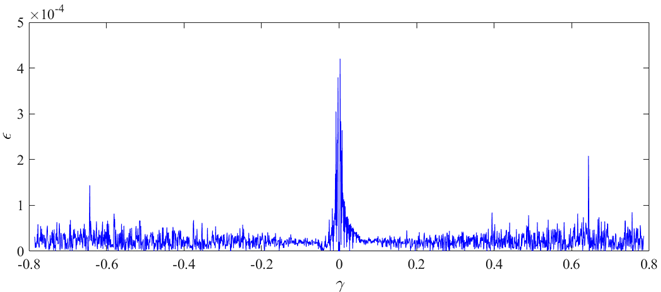}
	\caption{$\epsilon$ with respect to different  $\gamma$ ($\kappa=50$). }
	\label{fig.gt_gt_error}
\end{figure}

\begin{figure}[!t]
	\centering
	\includegraphics[width=1\textwidth]{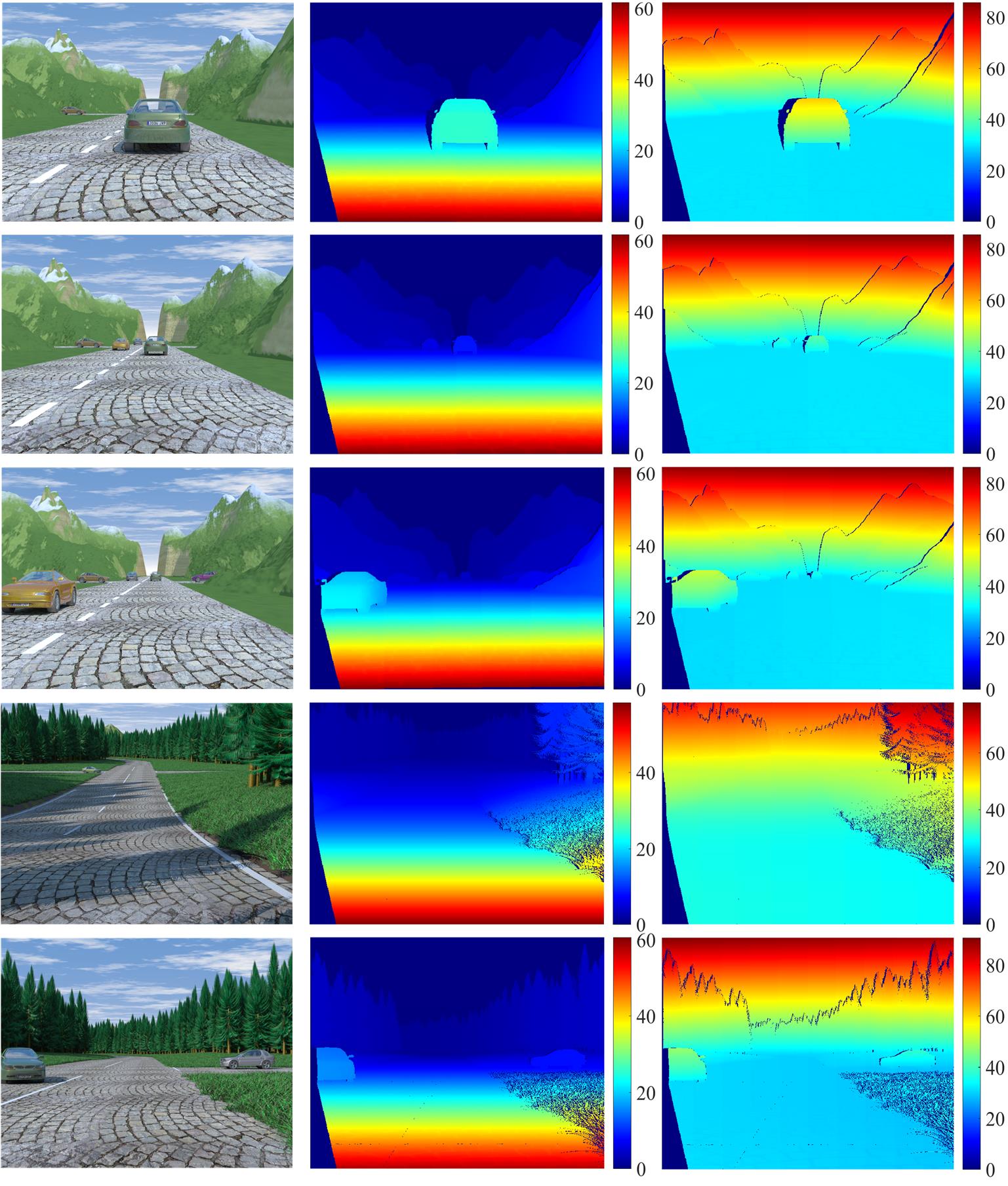}
	\caption{Experimental results of EISATS stereo datasets. }
	\label{fig.gamma_tran_result}
\end{figure}
To evaluate the accuracy of the proposed $\gamma
$ estimation algorithm, we create a ground truth image (resolution: $640\times480$), as shown in Figure \ref{fig.disp_gt1}, to simulate the disparity map of a non-flat road surface. The parabola of the road model is set to: $d(v)=100+0.3v+0.1v^2$. Then, the disparity map in Figure \ref{fig.disp_gt1} is rotated at different angles $\gamma$ between $-\pi/4$ and $\pi/4$. In each rotation, we estimate a roll angle $\tilde{\gamma}$ and compute the error $\epsilon=|\gamma-\tilde{\gamma}|$ between $\tilde{\gamma}$ and its ground truth $\gamma$. The maximum error is less than $3.7\times10^{-5}$ (around $0.0021^\circ$) and the average of the errors is around $2.3\times10^{-6}$ (around $0.0001^\circ$). Furthermore, Gaussian white noise $\kappa\omega$  is added to the created disparity map (see Figure \ref{fig.disp_gt2}), where $\omega$ is a random decimal number between $-1$ and $+1$, and $\kappa$ is scale parameter to control the intensity of the noise. In our experiments, $\kappa$ is set to $50$ to overestimate the noise. An example of the relationship between $\gamma$ and $\epsilon$ is shown in Figure \ref{fig.gt_gt_error}, where the average of the errors is approximately $0.0014^\circ$ (The maximum error is $0.0241^\circ$) and $\epsilon$ is relatively large when $\gamma$ is around $0^\circ$. This indicates that our proposed roll angle estimation can ensure high accuracy even when the disparity map is affected by noise.

We further evaluate the algorithm's accuracy using the stereo sequences from the EISATS database \cite{Vaudrey2008, Wedel2008}, where the roll angle is zero and some vehicles are on the road surface. Some examples of the experimental results are given in Figure \ref{fig.gamma_tran_result}. The disparity value of the road pixels become similar while the obstacle disparities change gradually. This occurs because the fitted disparity value $d(v)$ decreases gradually from the bottom of the disparity map to the top. However, when $v$ becomes smaller than the horizontal coordinate of the vanishing point, $d(v)$ becomes negative as proved in \cite{Ozgunalp2017}. The average of the roll angle errors $\epsilon$ for the EISATS database is approximately $0.0647^\circ$ which is still low.

\section{Conclusion}
\label{sec.pd_conclusion}

In this paper, a novel disparity transformation algorithm was proposed to extract road areas from dense disparity maps. By transforming the dense disparity map, the disparity value of the road pixels become very similar, while they differ greatly from those of obstacles and potholes. The roll angle $\gamma$ of a stereo rig can be estimated accurately by rotating the disparity map at different angles and identifying the angle which corresponds to the minimum fitting energy. To lower the computational complexity of $\gamma$ estimation, we utilised the golden section search to reduce the search range within the interval for each iteration. Rotating the disparity map around the estimated roll angle makes the disparity distribution of each row become very compact, which greatly improves the accuracy of the dynamic programming and the least squares fitting. Following the disparity transformation, the road area becomes highly distinguishable and can be extracted much easier.

\end{document}